\documentclass{article}
\usepackage{ijcai22}

\usepackage{times}
\usepackage{soul}
\usepackage{url}
\usepackage[hidelinks]{hyperref}
\usepackage[utf8]{inputenc}
\usepackage[small]{caption}
\usepackage{graphicx}
\usepackage{amsmath}
\usepackage{amsthm}
\usepackage{booktabs}
\usepackage{algorithm}
\usepackage{algorithmic}
\usepackage{url}

\title{SIAD: Self-supervised Image Anomaly Detection System}

\author{
Jiawei Li$^1$\and
Chenxi Lan$^1$\and
Xinyi Zhang$^2$\and
Bolin Jiang$^{1,2}$\and
Yuqiu Xie$^{1,2}$\\
Naiqi Li$^2$\and
Yan Liu$^1$\and
Yaowei Li$^1$\and
Enze Huo$^1$\and
Bin Chen$^3$
\affiliations
$^1$Manufacturing Department, Huawei Technologies, Dongguan, China\\
$^2$Tsinghua Shenzhen International Graduate School, Shenzhen, China\\
$^3$Harbin Institute of Technology, Shenzhen, China
\emails
li-jw15@tsinghua.org.cn
}
\begin{document}

\maketitle

\begin{abstract}
Recent trends in AIGC effectively boosted the application of visual inspection. However, most of the available systems work in a human-in-the-loop manner and can not provide long-term support to the online application. 
To make a step forward, this paper outlines an automatic image anomaly detection system called SIAD, working in a self-supervised learning manner, for continuously making the online visual inspection in the manufacturing automation scenarios. 
Benefit from the self-supervised learning, SIAD is effective to establish a visual inspection application for the whole life-cycle of manufacturing. In the early stage, with only the anomaly-free data, the unsupervised algorithms are adopted to process the pretext task and generate coarse labels for the following data. Then supervised algorithms are trained for the downstream task. With user-friendly web-based interfaces, SIAD is very convenient to integrate and deploy both of the unsupervised and supervised algorithms.  
\end{abstract}

\section{Introduction}
Manufacturing automation has rapidly advanced with the process of cloud computing and online visual inspection systems \cite{Amazon,Baidu,Huawei,Google}. By providing data labeling, GPU management, algorithm development, and cloud-to-edge deployment tools, current online systems follow canonical supervised learning manner and help to engage the human ability into the pipeline. However, since the online application continuously generates plenty of data, the manual operation is time-consuming and becomes a performance bottleneck. Therefore, there is a need to develop a less human involved long-term support system for the life-cycle of online applications \cite{Online}. 

\begin{figure}[htb]
\begin{minipage}[b]{1.0\linewidth}
  \centering
  \centerline{\includegraphics[width=8.5cm]{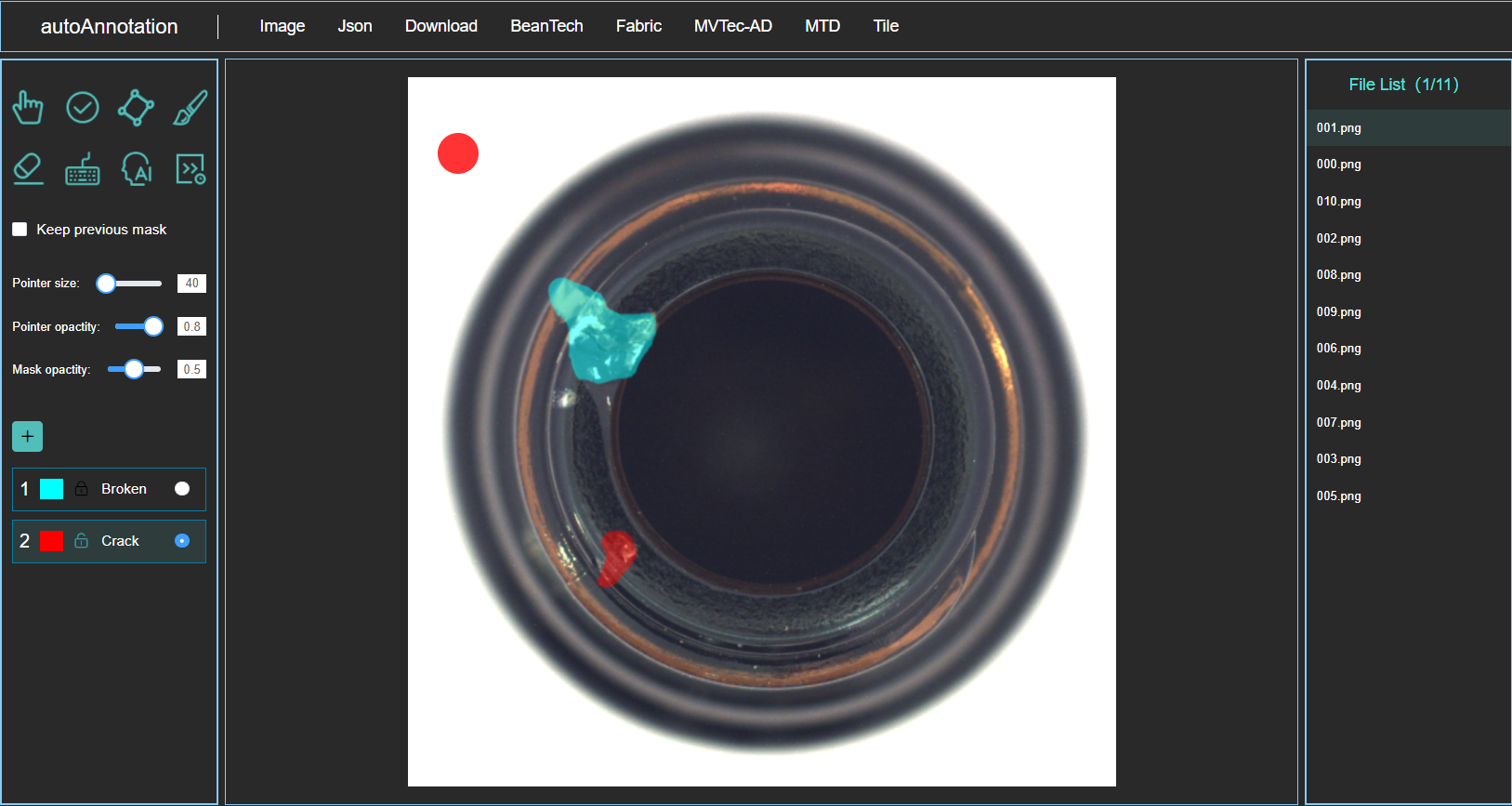}}
\end{minipage}
\caption{The web interface of our proposed SIAD system.}
\label{fig:web-interface}
\end{figure}

On the one hand, from the perspective of annotation, current supervised visual inspection relies on human expertise to make high-quality labels for the model training. 
However, with a human-in-the-loop manner, labeling image data by hand is a subjective task and inevitably to induce noise. 
Although some techniques about learning from noise labels \cite{NoiseLabel} are developed to relieve this problem, the performance is still limited by the generalization ability of the models.
Another way is to directly simplify human operations.  
In the medical image diagnosis scenario, some automatic systems are developed to assist the annotation and diagnosis of CT\&MRI images. For example, the web-based tool Medseg \cite{Medseg} and cross-platform software Pair \cite{Pair} provide painting brush, eraser, zoom-in, zoom-out, and other AI-driven auxiliary operations for the pixel-level modification, which requires less effort for introducing human knowledge than typical annotation tools (e.g., Labelme \cite{Labelme}). 

On the other hand, from an online learning perspective, current supervised visual inspection methods (e.g., YOLOv4 \cite{YOLOv4}, Faster-RCNN \cite{Faster-RCNN}) need to be trained with typical defect samples in advance. 
Because the real-world manufacture defects (such as scratch, stain, foreign material) usually have variable shapes and patterns, the few-shot learning and AutoML methods are adopted for tackling this kind of long-tailed detection problem \cite{Long_tail}. 
Although the few-shot learning does not require extensive training data and is useful for the tail defect classes, it is hard to tackle the open-set defects, which are never seen before \cite{F-S_O-S}. 
Besides, current AutoML techniques focus on algorithm selection and neural architecture search \cite{AutoML}. They can continue to learn good models from consistent data, but still have a gap to provide long-term support for online applications, especially when the incoming material and production are changed. 

In this paper, we propose a self-supervised image anomaly detection (SIAD) system to provide long-term support for online visual inspection and manufacturing automation. 
A user-friendly prototype system is developed \footnote{Link for trials: \url{http://139.9.56.97:8686/index.html}}
and shown in Figure \ref{fig:web-interface}. Powered by the object-oriented web front-end technologies \cite{FabricJS}, the generated annotations like geometrical shapes are regarded as separated canvas elements, thus easy to distinguish the semantic information of different defects. 
Due to the distinction and relationship of unsupervised proxy task and supervised down-stream task \cite{ContextEncoder,BYOL}, self-supervised learning has advantages for online learning. In our demo video \footnote{Demo video link: \url{http://139.9.56.97:8686/\#/video}},
an unsupervised anomaly detection model is trained on the image data of qualified products (do not need any label to indicate the defect), then predicts defects or generates pixel-level fine-graded labels for sustainable manufacturing. 

\section{The Prototype System}

\subsection{System Architecture}
Recently, Amazon launched the Lookout for Vision tool \cite{Amazon}, an anomaly detection solution that uses supervised learning to spot defects and anomalies on thousands of manufacturing images. It roughly has a typical two-stage workflow: creating labels for collected data and training a supervised model, then inferring the online data. Google also launched a new Visual Inspection AI tool \cite{Google} based on their cloud platform, which aims to help manufacturers to reduce defects. Due to the active learning and AutoML techniques, it achieves less labeling effort and higher training precision. However, it is still based on supervised learning and relies on human efforts for long-term support. 

\begin{figure}[htb]
\begin{minipage}[b]{1.0\linewidth}
  \centering
  \centerline{\includegraphics[width=8cm]{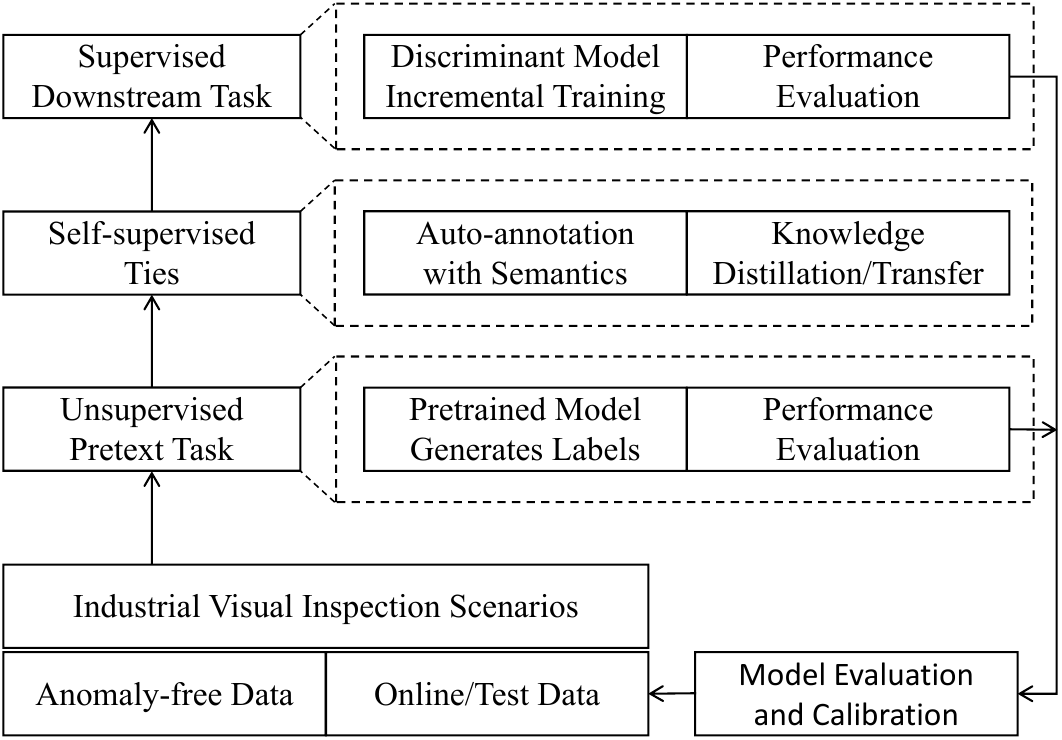}}
\end{minipage}
\caption{System architecture of self-supervised visual inspection.}
\label{fig:architecture}
\end{figure}

As shown in Figure \ref{fig:architecture}, our proposed SIAD prototype is designed with self-supervised learning, and has a closed-loop architecture to process the continuous online data with a user-friendly auto-annotation interface. 

\subsection{Manufacturing Automation}
Industrial visual inspection and medical imaging diagnosis have some common ground, while the former process manufacturing defects and the latter process visible diseases. 

The pandemic of Covid-19 speeds up automatic diagnosis techniques to some extent, especially when the medical resources are insufficient, AI-based tools and algorithms are developed to ease the shortage \cite{Diagnostics}. 
Medseg is an advanced web-based tool for computer-aided medical imaging diagnosis \cite{Medseg}. It follows a typical client-server architecture and can diagnose a single case at a time. The front-end page is written in vanilla JavaScript Canvas to support the pixel-level annotation. The back-end server provides 25 pre-trained AI models for download and uses the local GPU to make the automatic diagnosis.

Because the industrial manufacture highly scheduled on time and being uninterrupted, our proposed SIAD system is powered with some different techniques. It has an upgraded front-end written in the object-oriented Fabric.js Canvas, thus not only supporting the pixel-level annotation, but also requiring less effort and being faster when changing the semantic information of automatically generated annotations. 
Besides, the back-end of SIAD is powered by cloud computing and encapsulates the interface of different computing frameworks (e.g., Mindspore, Tensorflow, Pytorch, Halcon), thus is compatible with various hardware resources and algorithms. When the production environment is changed, the SIAD system requires less development and deployment costs. 

\subsection{Self-supervised Learning}
A typical self-supervised learning procedure has two stages. The pre-train stage has a pretext task to train the unsupervised model with unlabeled data. In the finetune stage, a downstream task is defined to train the supervised model with extracted features and labeled data. According to different pretext tasks and model architectures, the mainstream self-supervision can be summarized into contrastive, generative, or generative-contrastive based methods \cite{TKDE2021}. 

Due to highly explainable diagnosis results, the AutoEncoder and UNet models family are widely used in medical imaging and industrial manufacturing applications \cite{MIA2021,nnUNet,ToM2021}. 
To discriminate the open-set defects, a recent study tries to simultaneously perform the unsupervised image reconstruction and supervised anomaly detection tasks \cite{Dream}. It utilizes a Perlin noise generator to capture a variety of anomaly shapes, then joint train the AutoEncoder and UNet models on the augmented dataset. Different from relying on a noise prior, another approach trains a model to propagate the initial annotation in the pretext task and generates annotations for the downstream task \cite{Sli2Vol2021}. 

To design a prototype for online visual inspection, we leverage the incremental data for the self-supervised learning of open-set defects. The unsupervised image reconstruction and supervised image segmentation are defined as the pretext task and downstream task, respectively. A generative model is trained on the anomaly-free image and generates pixel-level labels for online data. 
Then train the discriminant model with the incremental data for sustainable defect detection.

\begin{figure*}[htb]
\begin{minipage}[b]{1.0\linewidth}
  \centering
  \centerline{\includegraphics[width=18cm]{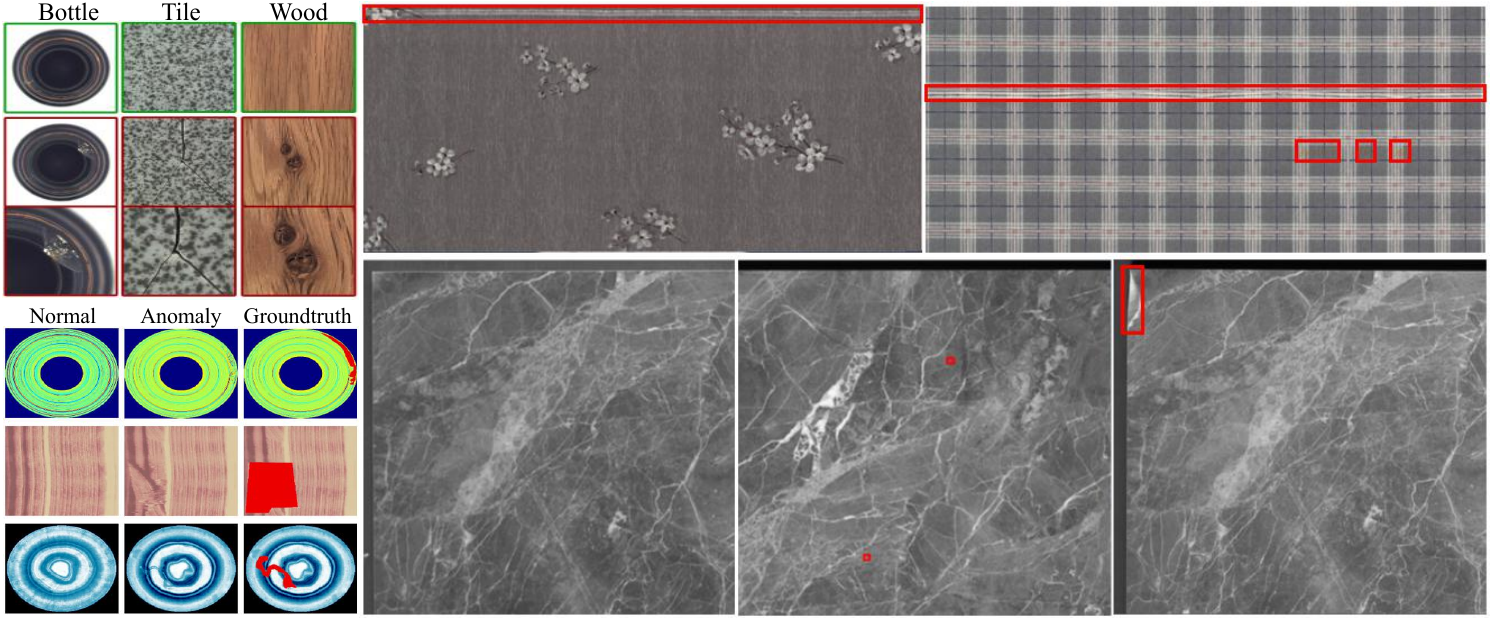}}
\end{minipage}
\caption{Four public available datasets for industrial visual inspection: MVTec-AD, BeanTech, TianChi-Fabric, and TianChi-Tile.}
\label{fig:dataset}
\end{figure*}

\section{Online Visual Inspection}
\subsection{Anomaly Detection Datasets}
MVTec-AD dataset \cite{MVTec-AD} contains 15 different textures or products. We design the demo routine with the \textit{Bottle} class, which is representative and usually used for the unsupervised anomaly detection task. It has a high-quality image with $900\times900$ resolution. The training set of the \textit{Bottle} contains 209 variable anomaly-free samples, while the testing set has 20 normal samples, 42 samples with broken defects, and 21 samples with contamination defects.

BeanTech dataset \cite{BeanTech} is a real-world anomaly detection dataset. It contains three different industrial products. The appearance of BeanTech Product 1 and 3 are similar to the MVTec-AD \textit{Bottle}. It is with $1600\times1600$ pixels and the anomaly-free samples are more variable. The training set has 400 anomaly-free samples, while the testing set has 21 normal samples and 49 not-good samples.

TianChi-Fabric \cite{Fabric} is a artificial fabric defect detection dataset. It contains more than 3,107 samples with object-level annotations for the supervised defect detection task. On the official annotation file, the dataset has a total of 16,457 defects that belong to 15 different types. 
Because the texture on different fabrics is quite different, this dataset also provides a standard anomaly-free template image for each kind of texture. Therefore, all the defect areas could be observed by a direct difference between the not-good samples and the corresponding template image. 

TianChi-Tile \cite{Tile} is a real-world tile defect detection dataset and was photographed from a fixed camera of a tile factory production line. It contains more than 12,000 samples with object-level annotations for the supervised defect detection task. Because some defects are very small ($<3$\% field of camera view), and there is no anomaly-free template image for some defective samples, this dataset is more difficult than the TianChi-Fabric dataset.

\subsection{Visual Inspection Routines}


According to the system architecture in Figure \ref{fig:architecture}, all the self-supervised visual inspection routines have three different stages. To simulate the data incoming of online applications, we independently and sequentially align the three stages to a train set, an online set, and a test set. 

The first routine in our prototype is with the MVTec-AD \textit{Bottle} and BeanTech datasets. We perform the unsupervised image reconstruction pretext task with a \textbf{training set}, which includes 10 anomaly-free samples. Then the pre-trained model is adopted to generate pixel-level labels for the \textbf{online set}, which includes 12 normal, broken, and contamination samples. After that, we perform a supervised image segmentation downstream task and evaluate the performance with a \textbf{test set}, which includes the other 12 defective samples. As shown in Figure \ref{fig:dataset}, the appearance of BeanTech Product 1 and 3 are similar to the \textit{Bottle}, thus we also have a routine to show the transfer learning ability in the prototype.

The second routine in our prototype is with the Tianchi-Fabric and Tianchi-Tile datasets. Following section 2.3 and not using the AutoEncoder and UNet, we further adopt a GAN inversion method \cite{DGP} to perform the downstream task, which has strong generalization ability and is capable for the productions with more variations.  

\section{Conclusion and Future Works}
In this paper, we design the SIAD prototype system for online visual inspection. Benefiting from self-supervised learning and a well-designed front-end page, we achieve long-term support for manufacturing automation and demo that. 

Now we can use the expert's experience on specific productions to make the auto-annotation with semantics. We plan to develop general automatic evaluation methods in the future, thus supporting some more real-life scenarios. 

\bibliographystyle{named}
\bibliography{iccv23_demo}

\end{document}